\documentclass[10pt]{article}
\usepackage{fancyhdr}
\usepackage[letterpaper]{geometry}
\usepackage{hicss}
\usepackage{times}
\usepackage[none]{hyphenat}
\usepackage{url}
\usepackage{latexsym}

\usepackage[frozencache,cachedir=minted-cache]{minted}

\usepackage{indentfirst}
\usepackage{graphicx}
\graphicspath{{images/}}
\usepackage[
    style=apa,
  ]{biblatex}
\usepackage{csquotes}
\usepackage{multirow}

\usepackage{adjustbox}
\usepackage[T1]{fontenc}

\fancypagestyle{empty}{%
  \fancyhf{} 
  \fancyfoot[C]{\textit{This paper has been accepted for the upcoming 57th Hawaii International Conference on System Sciences (HICSS-57)}}
  
}

\title{How well can machine-generated texts be identified and can language models be trained to avoid identification?}

\author{Sinclair Schneider\qquad Florian Steuber\qquad João A. G. Schneider\qquad Gabi Dreo Rodosek\\
Bundeswehr University Munich, Bavaria, Germany\\
\{Sinclair.Schneider, Florian.Steuber, Joao.Schneider, Gabi.Dreo\}@unibw.de}

\date{}

\begin{document}
\maketitle
\begin{abstract}
With the rise of generative pre-trained transformer models such as GPT-3, GPT-NeoX, or OPT, distinguishing human-generated texts from machine-generated ones has become important. We refined five separate language models to generate synthetic tweets, uncovering that shallow learning classification algorithms, like Naive Bayes, achieve detection accuracy between 0.6 and 0.8. 

Shallow learning classifiers differ from human-based detection, especially when using higher temperature values during text generation, resulting in a lower detection rate.
Humans prioritize linguistic acceptability, which tends to be higher at lower temperature values. In contrast, transformer-based classifiers have an accuracy of 0.9 and above. We found that using a reinforcement learning approach to refine our generative models can successfully evade BERT-based classifiers with a detection accuracy of 0.15 or less.
\end{abstract}

\subsubsection*{Keywords:}

Language Models, Language Model Detection, Transformer Reinforcement Learning

\section{Introduction}
Improving transformer models has led to the creation of higher-quality machine-generated text. This has led to the question of whether a distinction between human-written and machine-generated text is reliably possible.
The main concern is that it can be difficult for humans to make an accurate distinction because they focus on linguistic properties of the text rather than statistical features such as word probability, which are the focus of classification models.


According to \textcite{DBLP:conf/acl/IppolitoDCE20}, humans are better in their judgment if the number of unlikely words increases, whereas classification models exhibit the opposite behavior, prioritizing statistical evidence. Similar studies, including \textcite{DBLP:conf/acl/GehrmannSR19}, concluded that humans can detect fake texts with an accuracy of 54\%. 
Without computer-aided tools, humans achieve only a 50\% accuracy rate in identifying GPT-3 generated texts \parencite{DBLP:conf/acl/ClarkASHGS20}.

These findings highlight the need for a machine-guided decision process in reliably identifying artificially generated texts. This work, therefore, aims to investigate the accuracy and reliability of detection mechanisms when applied to GPT (Generative Pre-trained Transformer)-generated texts. 

We conducted a study that involved comparing different temperatures, sampling methods, and sample sizes using basic classification algorithms. The outcome revealed that these methods were insufficient in distinguishing between human-written and machine-generated texts. However, when more advanced classifier models like BERT were utilized, the results were more consistent and reliable.

Various approaches exist to trick or bypass these detection mechanisms, including paraphrasing the generated texts \parencite{DBLP:journals/corr/abs-2303-13408}. Introducing an alternative approach, we focus on transformer-based reinforcement learning to bypass the detection classifier. This technique was initially proposed by \textcite{DBLP:journals/corr/abs-1909-08593-neu} to fine-tune language models using human feedback. In contrast to the conventional training approach, we treat the feedback of the detector model as a reward during the reinforcement learning processing. Consequently, the generator model is rewarded most when the classifier incorrectly identifies the output as human-generated. We furthermore add additional linguistic constraints to refine the generated texts, avoiding exploitation of the classification model. This points out the limitations of detector models in detecting intentionally altered texts created by generative models that have been modified to evade the classifier by malicious actors.

\section{Related Work}

\subsection{Automatic Generation of Texts}

Different approaches have been used to create language models capable of generating texts. The most common architecture revolves around transformer models, including GPT and its predecessors, such as GPT-Neo-125M, GPT-Neo-1.3B, GPT-Neo-2.7B \parencite{gpt-neo}, GPT-J-6B \parencite{gpt-j}, OPT-125M, OPT-350M, OPT-1.3B, OPT-2.7B \parencite{zhang2022opt} and GPT-2 \parencite{radford2019language}. Further variants include Instruct-GPT \parencite{ouyang2022training} and T5 from Google \parencite{2020t5} model.


\subsection{Detection Techniques}
\textcite{orabi-2020} provide a further taxonomy of detection techniques, based on \textit{graphs} \parencite[e.g.][]{abou2020botchase}, \textit{crowdsourcing} \parencite[i.e. manually bot identification, e.g.][]{wang2012social}, \textit{anomalies} \parencite[e.g.][]{nomm2018unsupervised} and \textit{machine learning} \parencite{alothali-2018}.

However, since most of the approaches are either carried out by humans or make use of automatic tools, one can simplify them into two categories, i.e., \textit{human} vs. \textit{machine-based techniques}\footnote{this is true at least for all the mentioned exemplary studies, even though \textcite{orabi-2020} use different terminology, e.g. graph-based, anomaly-based, etc.}. Both of them are based on \textit{behavior} vs.  \textit{content} of the user. 
We focus on pure content, which is strictly speaking \textit{text} (in the case of Twitter posts, so-called \textit{tweets}).
\parencite{alothali-2018}

\subsubsection*{Human-based Detection.}

 When tasked to identify computer-generated documents, human individuals mainly focus on the semantic coherence of the presented texts. This contrasts machine-based detection approaches, which instead focus on statistical properties, such as word probabilities and sampling schemes \parencite{DBLP:conf/acl/IppolitoDCE20}. 
\textcite{DBLP:conf/emnlp/DuganIKC20} present a tool to evaluate human detection and conclude that it is relatively easy to fool humans for the above reasons.

\subsubsection*{Machine-based Detection.}

Statistical methods for machine-generated text detection are based on the fact that different modeling techniques leave detectable artifacts in the generated texts \parencite{DBLP:conf/acl/TayBZBMT20}. 

Furthermore, there exist various rule-based models to detect automatically generated texts, which are based on improbable word sequences and grammar \parencite{DBLP:journals/jasis/CabanacL21} or on similarity measures including word overlap \parencite{harada2021discrimination}. 

RoBERTa or BERT-based classifiers have downsides, such as their tendency to over-fit and the need to train a classification model every time a new generator is released. To solve this issue, zero-shot classifiers like DetectGPT \parencite{DBLP:journals/corr/abs-2301-11305} have been developed, requiring only a duplicate of the model that should be tested and a second language model to introduce random permutations to the test-string. Due to their principle of introducing random permutations to the test string, they cannot operate with a short input string.

To conclude, due to the novelty of most classification approaches, there are few scientific publications on bypassing them, like the paraphrasing-based one from \textcite{DBLP:journals/corr/abs-2303-13408}. Since we use reinforcement learning to adjust our generative model to bypass the classifier, our approach is based on the paper \enquote{Fine-Tuning Language Models from Human Preferences} by \textcite{DBLP:journals/corr/abs-1909-08593-neu}.



\section{Methodology}

Our methodology is structured as follows. We describe the data set used for and the training of generator models. Then, we investigate different approaches for machine-generated text detection, starting with shallow learning methods and ranging to transformer-based detectors. We present our reinforcement learning model aimed at generating tweets that evade previous detection approaches.

\subsection{Data set}

\begin{figure}[t]
\centering
\includegraphics[width=0.92\columnwidth]{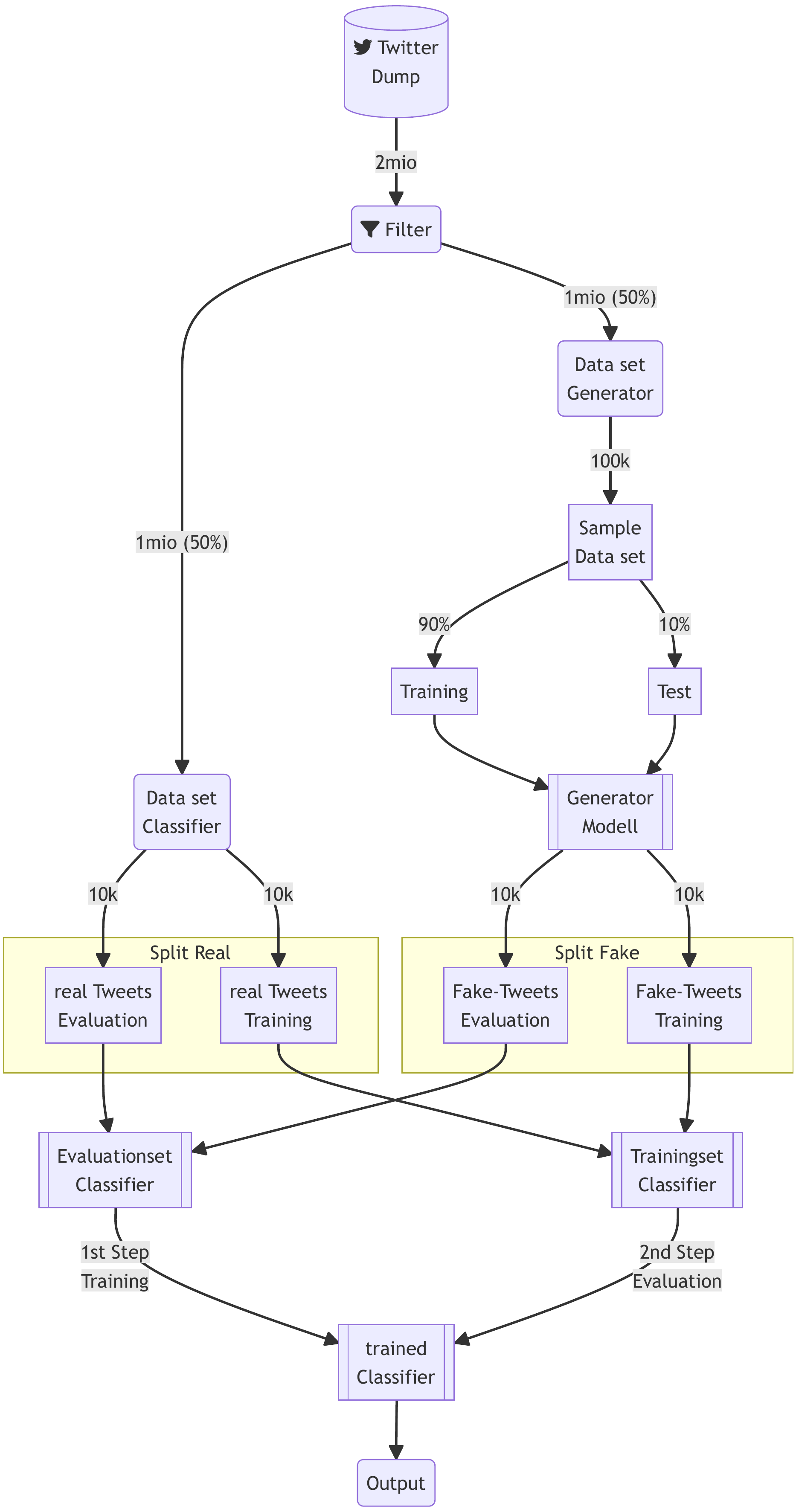}
\caption{\small 
Data pipeline used for modeling
}
\label{fig:flowchart}
\end{figure}

Our data set consists of tweets recorded between January and February 2020. The data is saturated with spam and advertising content, making it poorly suitable as direct training input.
We, therefore, applied a set of filter policies.

First, we filter tweets composed in English to achieve comparable results when evaluating against other state-of-the-art literature. Next, we extracted texts of verified users with less than $\geq100.000$ followers to avoid accounts that promote advertisements. Subsequently, we choose non-truncated tweets, as longer text might get truncated if not fully requested, which is undesirable for training. Furthermore, we restrict to original tweets and omit quotes, replies, and retweets to obtain a data set of more diverse texts. Finally, we check if the author of the tweets exhibits an average tweet volume (at most 20 tweets per day) to avoid spam content.

Overall, the data set consists of $k = 2.000.000$ tweets from $n = 136.450$ real accounts and is split into two equal parts, one for the transfer learning of the language models and one for testing the classifiers. We complement both training and evaluation data sets by adding real-world tweets. The generated tweets then form the two data sets \enquote{Fake-Tweets Evaluation} and \enquote{Fake-Tweets Training}. The counterparts (\enquote{Real Tweets Evaluation} and \enquote{Real Tweets Training}) are then built out of the original filtered tweets as illustrated in Figure \ref{fig:flowchart}.

\begin{table*}[t]
	\center
	\caption{Example tweets generated with different models with temperature = 1.0 and top-50 sampling}
     \begin{adjustbox}{width=1\linewidth}
    \small
	\begin{tabular}[t]{l|l}
	    \textbf{Model} & \textbf{Tweet} \\
        \hline
        GPT-Neo-125M & {\fontfamily{cmss}\selectfont
Kobe says new coronavirus warning on plane is too difficult to understand.} \\
        OPT-125M & {\fontfamily{cmss}\selectfont I'm sure a few will be added in a future update as part of the \enquote{Duke} legacy.} \\
        OPT-350M & {\fontfamily{cmss}\selectfont Good luck on the final stage of your tour!} \\
        GPT-Neo-1.3B & {\fontfamily{cmss}\selectfont The new album is out now; make sure you have the album download code for free.} \\
        OPT-1.3B & {\fontfamily{cmss}\selectfont Rangers' Henrik Lundqvist: \enquote{I'm not even thinking about' the trade rumors, says the goalie}} \\
        GPT-Neo-2.7B & {\fontfamily{cmss}\selectfont \#ValentinesDay: Today is the day to celebrate the greatness of yourself. And to appreciate\dots} \\
        OPT-2.7B & {\fontfamily{cmss}\selectfont A very cold, chilly \#day for \#Lincoln and \#Omaha \#MorningWeather}\\
        GPT-J-6B & {\fontfamily{cmss}\selectfont\enquote{This is how we play games!} Let’s hear \enquote{The Box} tonight with @OzzyOsbourne\dots} \\
        
	\end{tabular}
     \end{adjustbox}
	\label{tab:methodology-example_tweets}
\end{table*}

\subsection{Training of Generative Models}\label{sec:transferLearning}

Our initial focus is to examine the efficacy of a fine-tuned LLM when generating tweets, 
as we believe they are more challenging to create than texts containing longer passages. Additionally, we aim to assess the capability of various existing detectors in identifying machine-generated short texts. Our experiments range from simple methods such as Naive Bayes to more advanced models such as BERT.

As a basis for our research, we make use of various GPT variants, including GPT-2 \parencite{radford2019language}, GPT-J-6B \parencite{gpt-j}, GPT-Neo-125M, GPT-Neo-1.3B, GPT-Neo-2.7B \parencite{gpt-neo}, OPT-125M, OPT-350M, OPT-1.3B and OPT-2.7B \parencite{zhang2022opt}. 
Despite the availability of larger free models, we opt to utilize transfer learning-compatible models on a machine equipped with an A6000 GPU and 128GB of RAM.

Each fine-tuned model generates two sets of fake tweets with a given sampling strategy and temperature. These are used to train and evaluate the detector model. The size of the generated sets varies, ranging from 10.000 for comparisons of simple Bag-of-Words classifiers to 100.000 for training BERT classifiers. Examples of generated tweets can be taken from Table \ref{tab:methodology-example_tweets}.

\subsection{Shallow Detectors for machine-generated Texts}
We begin by exploring the effectiveness of a Naive Bayes classifier using Bag-of-Words (BoW) feature inputs to identify synthetic tweets. This classifier is chosen due to its simplicity and short training time. Due to the restricted number of features that BoW extracts from the texts, modifications in the generation parameters of language models, such as sampling methods, \textit{k}-values, \textit{p}-values, etc., exert a significantly greater influence on the classifier's output. 

We aim to examine how the parameters of the generator, namely temperature, sampling strategy, sampling size, and model size, influence the accuracy of the classifier. For this, we performed a grid-based approach in training the classifier with different parameterizations, each time conducted using 10,000 real and generated tweets.




\subsubsection*{Temperature.}

A generator's temperature $\tau$ alters the probability distribution of the next word. A low value sharpens the distribution curve, resulting in more accurate sentences. In contrast, a higher value flattens the curve and thus allows a higher variety of words to be chosen, increasing the outcome's variance at the cost of linguistic coherence. In our experiments, we vary temperature values ranging from 0.8 to 1.4.



\subsubsection*{Sampling strategy.}

LLMs produce sentences by generating one token at a time. Token selection depends on the selected sampling method. For example, greedy search selects the token with the highest probability as the predecessor, whereas random sampling strategies pick the next token based on its probability. This process can be limited to a certain number (k-sampling) or a combined probability (p-sampling) of the most probable next tokens. Furthermore, to enhance the unpredictability of word generation, one can employ the typical sampling technique using conditional entropy
\parencite{meister2022locally}.

\subsubsection*{Sampling size.}

It is well known that both shallow and deep machine learning models yield better results when trained on larger data sets. This improvement diminishes with a certain amount of data depending on the complexity of the model. 
We test each classification algorithm with a training set of varying sizes (1k, 10k, 50k, and 100k tokens) to investigate the correlation and saturation.

\subsubsection*{Model size.}

OpenAI's GPT family has achieved notable language modeling advancements through substantial parameter size increases. While larger amounts are expected to generate better text results, they are more effective in creating longer texts. This is due to the larger context window associated with more complex models. Since our experiment involves generating short tweets, we aim to investigate if the model size affects the detectability of the language model. Our study compares models ranging from 125 million to 6 billion parameters.

\subsection{Transformer-based Detectors for machine-generated Texts}\label{sec:transformerBasedDetectors}

Given their state-of-the-art capabilities in text classification, transformer-based detectors are more likely to be used in production. We opted to use BERT as the primary reference model in the transformer family. Its performance is nearly on par with more advanced versions such as RoBERTa, as assessed by the MultiNLI benchmark \parencite{MultiNLI}. In their evaluation, BERT-Large records a score of 88\% \parencite{BERT-Large}, RoBERTa at 90.8\% \parencite{RoBERTa}, and DeBERTa at 91.1\% \parencite{DeBERTa}.

The evaluation process is similar to the shallow detectors, except that BERT's embeddings replace the BoW representation. Because of the training complexity, we refrain from fitting multiple BERT models with varying parameters.


\subsection{Reinforcement Learning to bypass the Detector}\label{sec:ReinforcementLearning}

Our final contribution shows that even an advanced detection model such as BERT can be bypassed. We utilize a reinforcement learning approach for text generation that progressively learns from the detector output of previously generated texts.
This approach roughly consists of three steps: rollout, evaluation, and optimization \parencite{vonwerra2022trl}, as discussed below. 

The first step includes the model's rollout. Here, the language models from Section \ref{sec:transferLearning} generate an artificial tweet. For this, a model is provided with the initial part of an original tweet and is required to finish the sentence. In addition to completing tweets, it sometimes has to generate full texts independently to prevent overfitting on short texts.


Secondly, the evaluation step takes place. Texts and responses are supplied to the corresponding BERT classifier, as described in Section \ref{sec:transformerBasedDetectors}. Should a classifier identify the text as composed by humans, a positive reward is assigned to the reinforcement learning algorithm, otherwise a negative one. Raw logits tend to perform best in this case. 

Finally, the optimization step takes place. The log probabilities of the tokens are calculated to compare the active language model to the reference model. This is part of the reinforcement learning described by \textcite{DBLP:journals/corr/abs-1909-08593-neu} and prevents the adjusted model from overoptimizing.


We now delve into more detail about our proposed handcrafted reward function. This reward supplements the detector's response and imposes a separate penalty on generated text, even if classified as human-written, in cases where specific linguistic constraints are not satisfied. Calculating rewards is depicted in Figure \ref{fig:rewardcalculation}. 
If one or more of the supplementary rules listed below produce unfavorable results, the smallest one is chosen. Conversely, if all the additional rules yield favorable results, the value indicating that humans created the text is selected.


\begin{figure}[t]
\centering
\includegraphics[width=0.99\columnwidth]{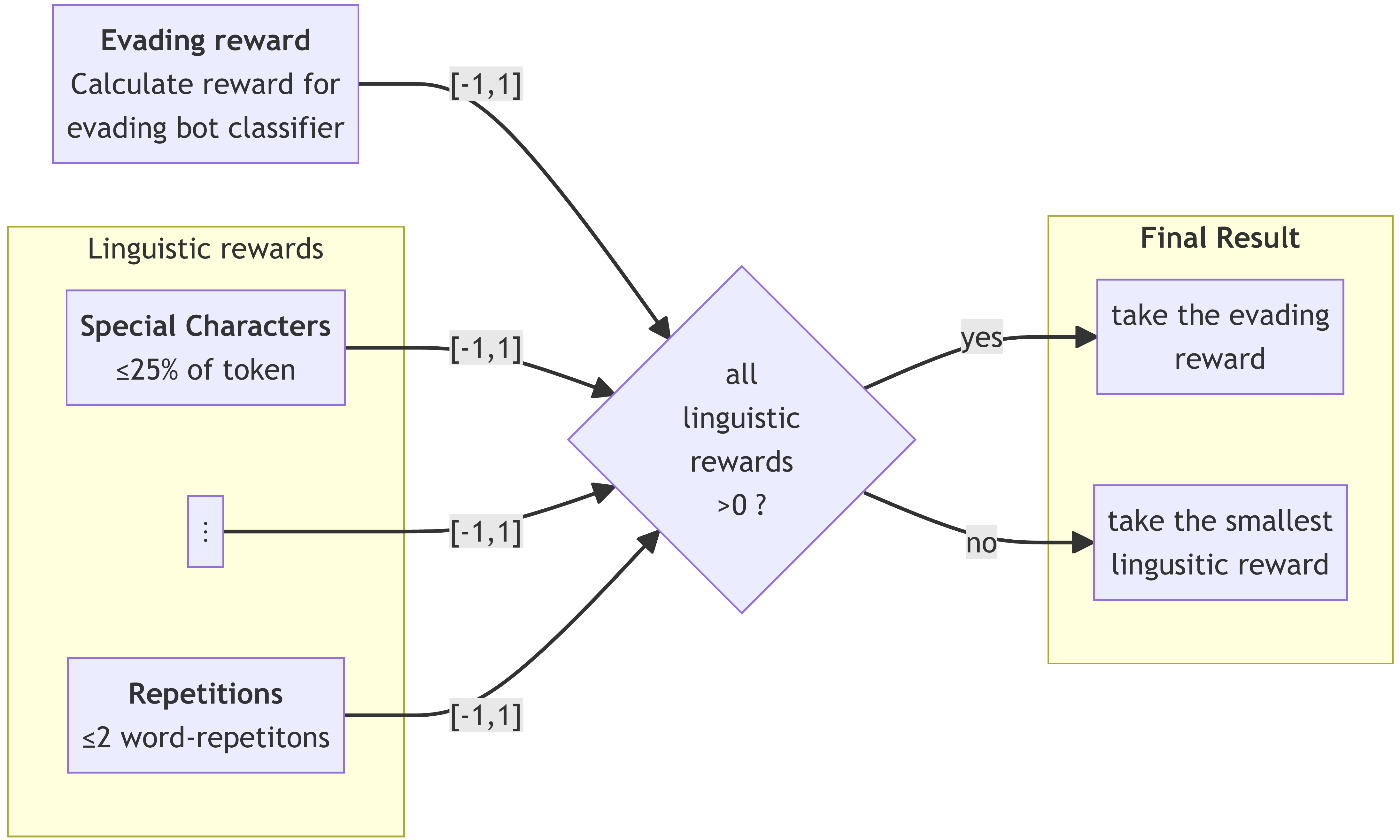}
\caption{\small 
RL reward calculation
}
\label{fig:rewardcalculation}
\end{figure}

\subsubsection*{Special characters.}

A text should not include more than 25\% of special characters. If it does, the model assigns a linearly decreasing negative reward of up to -1 if the text consists exclusively of special characters.

\subsubsection*{Repetitions.}

If a text contains more than two repetitions of the same word, it is assigned a negative reward of up to -1 when repeated eight times.

\subsubsection*{Linguistic acceptability.}

The linguistic acceptability is checked by the usage of a DeBERTaV3-v3-large classifier \parencite{he2021debertav3} trained on the Corpus of Linguistic Acceptability \parencite[CoLA;][]{warstadt2018neural}. Since this CoLA data set contains sentences marked as acceptable or unacceptable by humans from a grammatical point of view, the trained DeBERTa model is able to judge how good or acceptable a given sentence is. If the value is below a threshold of 40\%, a negative reward is returned, which minimizes to -1 at 0\% acceptability. Note that we relax the threshold value to 30\% when applied to models with over 2 billion parameters, as they are harder to train. 

The thresholds mentioned are not taken over from a reference paper but are determined empirically. Generally, it is recommended to aim for a higher threshold as it directly affects the text's linguistic acceptability. Nevertheless, we discovered that setting the threshold too restrictively can lead to the failure of the reinforcement learning process. In such cases, the generated text might be classified as non-human, rendering it impossible for the model to obtain rewards and impeding its learning progress.

\subsubsection*{Dictionary.}

To obtain natural texts, at least 25\% of tokens have to appear in a dictionary. Otherwise, a linearly increasing negative reward is assigned, which rises to -1 if none of the words appear in a dictionary. We remark that this value should be chosen relatively low in the context of tweets, as short texts typically contain more abbreviations and slang language.

\subsubsection*{Word Emoji relationship.}

A tweet must not contain more emojis than words. If it does, the reward is progressively penalized, up to a maximum, if only 25\% of the tweet is formed by words. Emojis are not counted as special characters. 

\subsubsection*{Number of Emojis.}

There should not be more than three emojis in one tweet. Every additional Emoji leads to a negative reward of -0.4 up to -1.

\subsubsection*{Repetition of the Query.}

To generate unique texts different from the input query, we assign a negative reward if more than half of the query is included in the output. This reward decreases to -1 if the entire query is repeated.

\subsubsection*{Special Token.}

There should not be more than two special tokens in one generated tweet, including BOS (beginning of the sentence) or EOS (end of sentence) delimiters. Every additional token gets a negative reward of 0.4 up to -1.

\begin{figure}[t]
\centering
\includegraphics[width=0.8\columnwidth]{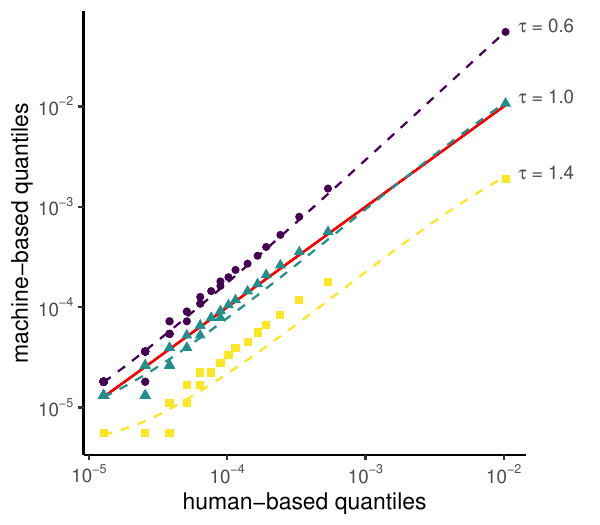}
\caption{\small Humans' against machines' word probability distributions
}
\label{fig:qqplotone}
\end{figure}

\subsubsection*{Same start.}

Sometimes, we require the model to generate tweets without an input query. In such cases, it is essential that the model still generates diverse outputs. Consequently, a negative reward is returned if more than 10\% of the tweets within a single training batch start with the same word. This value again decreases linearly to -1 if 20\% of all sentences have similar starts.

\subsubsection*{Numbers at the start.}

Like the previous policy, we do not want tweets to start with numbers frequently. Therefore, if the model grasps that a classifier can be circumvented by a sentence beginning with a number once, it may also learn that a similar sentence, or one with the same pattern, could be effective on another occasion. Hence, only 10\% of the sentences within a training batch are allowed to start with a number, resulting in a negative reward of up to -1 if there are more.

\subsubsection*{Unknown characters.}

Occasionally, language models may insert filler characters, represented as unknown characters. This phenomenon typically occurs when unknown characters are included in the data set used for fine-tuning. To prevent the model from using these characters to bypass the classifier, generating them will result in a negative reward starting with -0.5 and decreasing if the replacement character appears more than once.

\vspace{1em}
These additional optimization rules have been found by analyzing various preliminary runs and observing the logs for requests and responses during RL training. It is worth noting that RL can function without these rules, but the results are considerably inferior, as demonstrated by outcomes such as: \enquote{Something for Administrator930 Macy’s Displays! RIP Family Members}. In this particular case, the rule for linguistic acceptability would have avoided a positive reward. 

For bigger models, the reward of the bot classifier can be multiplied by a factor, e.g., 10, to give the task of bypassing the classifier a higher priority than having excellent sentences.

\subsection{Applicability to other text-domains}\label{sec:Applicability}
We conducted a second iteration using a CNN and Daily Mail data set \parencite{cnnDailyMailDataset} to demonstrate the practical application of the reinforcement learning approach. This also proved that our approach can be adjusted to generate fake news. The training and testing procedure for this iteration was similar to the fake tweets, except the linguistic refinement filters were reduced to only one. This filter ensured that not all generated texts start the same.
\begin{figure}
\centering
\includegraphics[width=1\columnwidth]{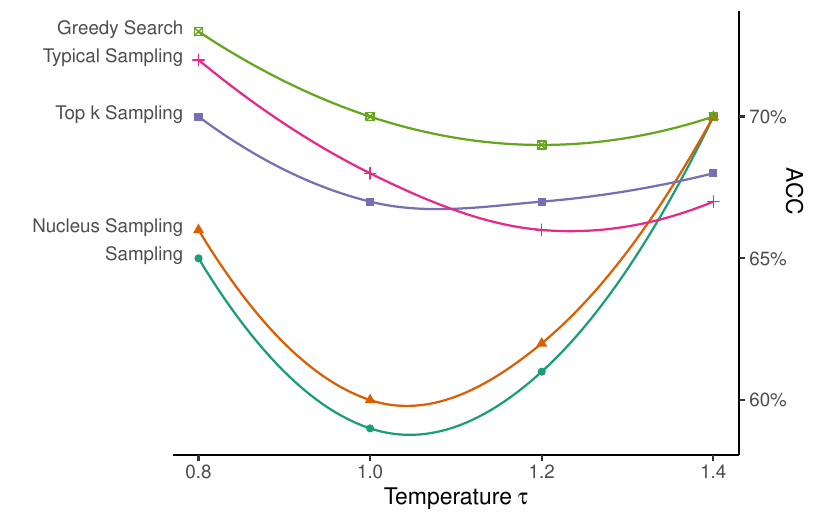}
\caption{\small 
Sampling methods' detection rates by temperature (top k = 100, nucleus p = 0.95)
}
\label{fig:samplingMethods}
\end{figure}

\section{Results}


\subsection{Evaluation of Shallow Detectors}
\subsubsection*{Word Distributions.}


By altering the probability distributions of the generating language model, statistical-based classifiers like Naive Bayes can detect changes more easily. In Figure \ref{fig:qqplotone}, a quantile-quantile plot visualizes the statistical differences that these classifiers rely on. The plot shows the probability distributions of words present in a corpus with different temperature values, which differ significantly from the word distributions in human-written texts.


This shows how machine-based detection differs significantly from human-based detection. A classifier can detect changes in the probability distribution in both directions, while humans rely on language comprehension. The level of linguistic acceptability increases when the distribution curve is sharpened and the probability of the most likely next words increases. However, reducing the temperature to achieve this comes at the expense of reducing the information content.


\begin{figure}[ht]
\centering
\includegraphics[width=0.95\columnwidth]{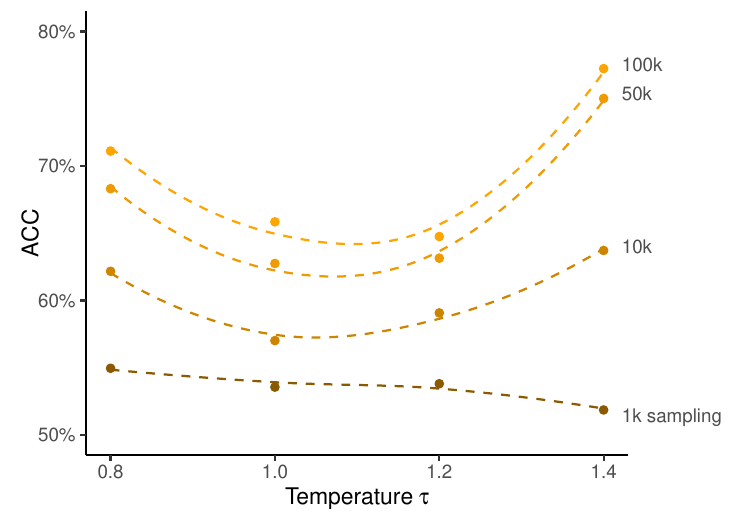}
\caption{\small Comparison of different sampling sizes}
\label{fig:sampling1to100k}
\end{figure}

\subsubsection*{Sampling Schemes.}

Next, we compare how different sampling methods used for generating texts affect the detection rates of shallow models. The results can be seen in
Figure \ref{fig:samplingMethods}. The easiest method to detect is greedy search, selecting the most likely next word based on maximum likelihood. Typical-p sampling ranks second and relies on conditional entropy to deviate from the original distribution. Top-k sampling also differs significantly from the initial distribution as it does not consider the distribution curvature. Finally, nucleus sampling, which considers the various shapes of the probability curves for the next token, is most evasive after sampling using pure randomness.


\subsubsection*{Sampling Size.}

Shallow learning models, including Naive Bayes, require small amounts of training data compared to deep learning models such as BERT. In Figure \ref{fig:sampling1to100k}, we have listed how different training data sizes affect a shallow detector's accuracy in successfully identifying machine-generated texts. As can be seen, even with a training batch consisting of as little as 1000 data points, the detector can still capture some dependencies. Increasing the size of the training data improves the detector's quality. However, the improvements successively diminish above a certain threshold, from 50k to 100k.

\begin{figure}
\centering
\includegraphics[width=1\columnwidth]{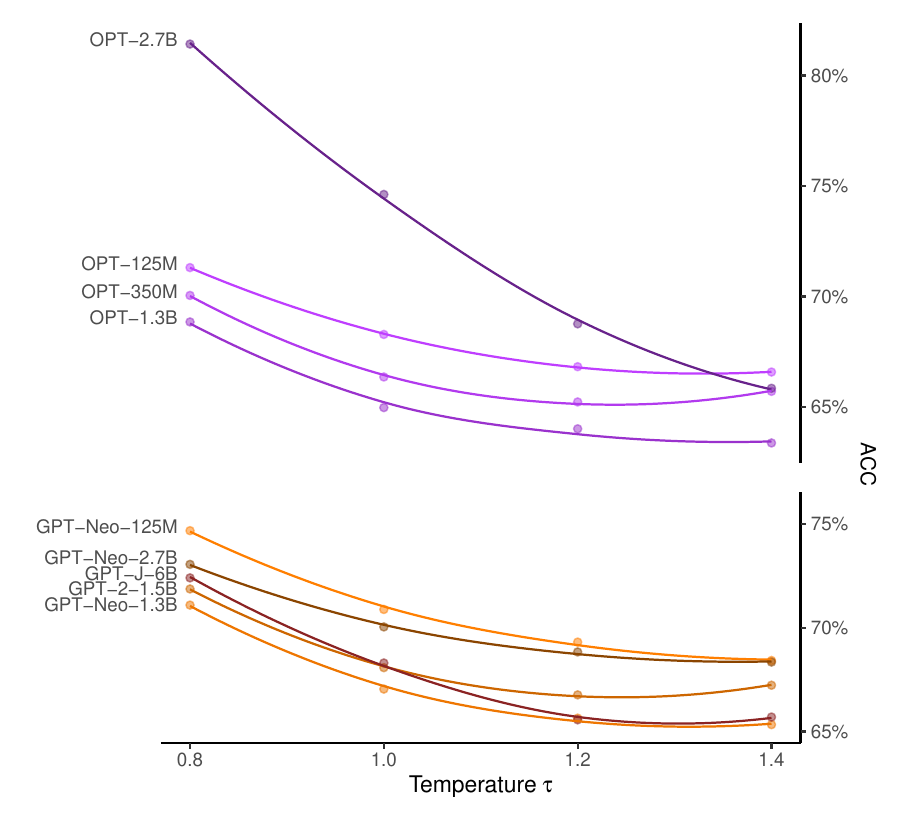}
\caption{\small Generator models' ACC by temperatures
}
\label{fig:modelcomparison}
\end{figure}

Contrary to the sample size, the parameter size of the evaluated models has a minor impact on the classification process, as seen in Figure \ref{fig:modelcomparison}. In this figure, we have grouped GPT and OPT model families in similar colors, where darker lines represent models with more internal parameters. One possible explanation for this behavior is that even though bigger models typically perform better on larger text passages due to larger context windows, the generation of short texts hardly benefits from this property. In addition, given that the generator family (OPT vs. GPT) leads to more significant differences in classification accuracy, we hypothesize that model design and training corpus inherently play a more crucial role in generating short texts than sheer model size.

\subsection{Evaluation of Transformer-based Detectors}

\begin{table}[t]
        \begin{center}
        \caption{F1-Scores before and after RL}
        \label{tab:BERTcomparison}
            \begin{tabular}{llcc} 
            &&\multicolumn{2}{c}{\textbf{BERT F1-Scores}}\\\cline{3-4}
            \textbf{Model}  & \textbf{parameter} & \textbf{\textit{pre}} & \textbf{\textit{post}}\\\hline 
            GPT-Neo &  125M & $0.952$ & $0.031$ \\
            GPT-Neo &  1.3B & $0.933$ & $0.146$\\
            GPT2    &  1.5B & $0.929$ & $0.090$\\
            GPT-Neo & 2.7B  & $0.947$ & $0.004$ \\
            GPT-Neo* & 2.7B  & $0.959$ & $0.173$ \\
            \multicolumn{4}{l}{\footnotesize{*Fake news based on CNN and Daily Mail data set and model}} \\
            \end{tabular}
        \end{center}
\end{table}

\begin{table*}[t]
	\center
	\caption{Example of the reinforcement learning process}
	\begin{tabular}[t]{l|l|r}
	    \textbf{query} & \textbf{response} & \textbf{reward} \\
        \hline
        FINAL & {\fontfamily{cmss}\selectfont UPDATE: Chancellor says he has not shown anxiety} & $0.8019$\, \\
        {\fontfamily{cmss}\selectfont\textless |startoftext|\textgreater} & {\fontfamily{cmss}\selectfont see you this weekend! we'll be here again cider is back in v} & $0.9674$\, \\
        {\fontfamily{cmss}\selectfont\textless |startoftext|\textgreater} & {\fontfamily{cmss}\selectfont flush those lemons. This food ain’t good} & $0.9628$\, \\
        Ozzy Osbourne cancel & {\fontfamily{cmss}\selectfont s gig at studio theatre tonight over coronavirus 2020 forced} & $-0.0089$\, \\
        Today we'll know & {\fontfamily{cmss}\selectfont soon morning if there's blood somewhere in} & $-0.1756$\, \\
        The best \#adventure this & {\fontfamily{cmss}\selectfont year on 2020, dudes. Can we really} & $-0.2899$\, \\
        {\fontfamily{cmss}\selectfont\textless |startoftext|\textgreater} & {\fontfamily{cmss}\selectfont Yeah it’s weird how local commentators} & $0.9696$\, \\
        
	\end{tabular}
	\label{tab:ReinforcementLEarning}
\end{table*}

The introduction of self-attention layers in transformer-based models has led to significant improvement in their performance. These layers allow the classifier to translate a token into vector space based on the surrounding tokens. As a result, all evaluated BERT models have achieved an F1-Score of over 0.9 (cf. Tab.~\ref{tab:BERTcomparison}, column \textit{pre}). The parameters for tweet generation included top 50 sampling with a temperature of $\tau=1.0$. While these results are impressive, there is a possibility that the BERT classifier might overfit one specific method and perform poorly on other generative models.

\subsection{Evaluation of Reinforcement Learning}

Transformer-based classifiers are generally very reliable in distinguishing real and generated content, as investigated in the previous sections. However, we show that a carefully fine-tuned reinforcement learning approach can bypass these well-performing models. To obtain reliable results, much emphasis has to be given to optimizing the model's hyperparameters, including learning rate, mini-batch size, chosen optimizer, and threshold for detecting linguistic acceptability. Based on empirical analysis, experiments have shown that a learning rate of $5\cdot10^{-5}$ and a mini-batch size of 4 tend to yield the best results.

For the GPT-Neo-2.7B model, we reduced the linguistic acceptability threshold from 0.4 to 0.3 since bigger models are more susceptible to manual interventions. Similarly, we exchanged the Adam optimizer within the 2.7 billion parameter model with the novel Lion optimizer \parencite{https://doi.org/10.48550/arxiv.2302.06675}, reportedly outperforming the first one.


We shared the same sampling technique during the reinforcement learning and evaluation phases to ensure reliable results. Table~\ref{tab:ReinforcementLEarning} presents a snippet of the log that documents queries, responses, and rewards during the reinforcement learning process. As stated earlier, a positive reward is only possible if both the query and response have been classified as human and none of the supporting rules outlined in section \ref{sec:ReinforcementLearning} have yielded negative results. In this case, we solely rely on the BERT classifier's reward to detect generated text. 

We list the classifier's F1-scores before and after the reinforcement learning process in Table ~\ref{tab:BERTcomparison}.
The 2.7 billion model had a lowered threshold for linguistic acceptability, resulting in significantly better results, but at the expense of lower linguistic quality. The experiment adapted to train the generator using CNN and the Daily Mail data set demonstrates the applicability of our approach to other text domains. Our experiment with four open-source models showed that a BERT classifier can be bypassed using a reinforcement learning training method.
\section{Conclusions and Outlook}

In this paper, we highlighted the significance of sampling techniques and the essential role of employing advanced models like transformers to detect generated texts accurately. 
We also demonstrated how a malicious actor could adapt generative models to evade a detector if accessible. However, this process is not simple and requires substantial computational power and hyperparameter tuning. As the parameter count of the generative model increases, the reinforcement learning process becomes more complex, and the output quality may decrease. While the risk of malicious actors modifying language models remains more theoretical than practical, it retains a degree of plausibility.

In various scenarios, such as completing homework in schools, submitting assignments at universities, or identifying suspicious campaigns on social media platforms like Twitter, it is essential to differentiate between human-written and machine-generated texts. OpenAI previously provided a detection model for this purpose, but it was discontinued in July 2023 due to inconsistency in distinguishing between the two \parencite{OpenAI}. Our experiments have revealed a trade-off between creating easily identifiable texts that lack diversity and generating more diverse texts that are harder to differentiate but might not be as readable to humans. Therefore, larger AI-generated models may be more challenging to spot as they can combine both linguistic acceptability and diversity, thanks to their size and thus the ability to generate a higher range of plausible following tokens while still producing high-quality texts. Consequently, the trade-off between high-quality texts and low detectability is expected to diminish.

Therefore, it is essential to consider that language models may become undetectable. In such a scenario, students using ChatGPT to complete their homework could be viewed as a lesser evil. This could lead to the creation of customized social engineering bots that steal personal data without human input from the attacker, the development of malicious code by those without coding knowledge, and customized spear phishing emails.


Based on promising approaches such as DetectGPT \parencite{DBLP:journals/corr/abs-2301-11305} but also the withdrawal of the OpenAI classification model and our findings, we would like to emphasize that detecting and preventing these types of incidents is an emerging research field with ample opportunities for further publications.

\section*{Acknowledgements}
The authors would like to thank the Chair for Communication Systems and Network Security for their valuable discussions and feedback as well as the Research Institute CODE for providing hardware.




\printbibliography

\end{document}